# Contributed chapter to "HANDBOOK OF RESEARCH ON ARTIFICIAL INTELLIGENCE, INNOVATION AND ENTREPRENEURSHIP" to be published by Edward Elgar Publishing

*Loizos Michael*
Open University of Cyprus &
CYENS Center of Excellence
***loizos@ouc.ac.cy***

**Abstract:** This chapter discusses AI from the prism of an automated process for the organization of data, and exemplifies the role that explainability has to play in moving from the current generation of AI systems to the next one, where the role of humans is lifted from that of data annotators working for the AI systems to that of collaborators working with the AI systems.

**Acknowledgements:** This work was supported by funding from the EU's Horizon 2020 Research and Innovation Programme under grant agreements no. 739578 and no. 823783, and from the Government of the Republic of Cyprus through the Deputy Ministry of Research, Innovation, and Digital Policy.

# Table of Contents

# Explainability and the Fourth AI Revolution

Explainable automated organization of data. Although admittedly not a comprehensive definition of the wide scope of Artificial Intelligence (AI), this phrase does capture how AI has come to be viewed in terms of how it has driven innovation, especially in the recent past, and how the foreseen potential dangers of uncontrolled automation have led to a strong socio-scientific movement in support of explainability.

Having begun as a scientifically-driven quest to understand the multiple and diverse facets of human intelligence to a degree that would allow for their replication in a mechanical manner (McCarthy, et al., 1955), AI has evolved over the past decade into a mostly engineering quest to develop systems that solve specific and specialized problems, retaining perhaps some inspiration from human intelligence and the brain, but placing emphasis on the performance of those systems in certain testing conditions rather than on the principled understanding of their operation and their compatibility with human cognition.

Facilitated by the vast amounts of available data — themselves a consequence of the disruption being brought about by the World Wide Web and the Internet of Things — AI has come to be treated as a synonym, at least in the eyes of the general public, of a technology that crunches data in some unspecified manner, and identifies hidden knowledge that is encoded in the data in a manner that can be usefully applied in novel contexts to make decisions. This view might be more pronounced in the innovation and entrepreneurship world, where terms like "powered by AI" are being used as buzzwords to characterize any such data-driven decision-making system for some specialized task of interest.

The automated acquisition and management of knowledge supported by AI is itself already disrupting innovation (Manyika, et al., 2013) and affecting the daily lives of people, with or without their consent, so that relevant stakeholders have started considering the potential downside of the ongoing disruption. Requiring AI systems to explain how they automatically organize data is currently taking center stage across the quadruple helix, as one way to alleviate any foreseen downside of this AI-driven disruption.

The explainability requirement on AI systems is, on the one hand, a natural effect of the pendulum-like process of scientific discovery that took place in the short history of AI. The early days of AI saw an emphasis on developing systems that manipulated symbols, followed by work on neural (subsymbolic) representations, a subsequent return to expert systems, and the more recent emphasis on deep neural representations. As researchers pushed the limits, and identified the weaknesses, of approaches at either extreme of the pendulum swing, they sought to overcome those limitations by exploring alternative paths towards the goal of developing AI systems, ending at the other extreme. Explainability can be seen, at least to some extent, as yet another swing of the pendulum towards the use of symbols.

A second reading of the emergence of explainability in AI has a socio-technical flavor. The society is realizing all the more clearly that AI systems that have been developed in the past decade, during the last swing of the pendulum, are not based on mature technologies, but on ones still in their "teenage years". The initial overconfidence brought about by the advent of deep learning is not unlike the overconfidence exhibited by a teenager compared to their actual competencies, as they pass from the low-competence low-confidence age of a baby to the high-competence high-confidence age of an adult (Kruger & Dunning, 1999). The shift towards explainability is but a somber realization that AI systems need to climb down the so-called "mountain of stupidity" and into the "valley of humility" before they acquire competencies that match the confidence that can be reasonably ascribed to them.

In this context, explainability acts as a tool towards the certification of AI systems, not only in terms of their functional performance against some objective metric of accuracy as typically done to date, but more critically in terms of the cognitive compatibility of their reasoning against the subjective needs and abilities of the particular group of humans for which an AI system has been designed and developed. Consequently, evaluating the quality of explanations and being mindful of the extent to which each type of explanation can contribute to this certification process, takes a center stage in designing AI systems.

This chapter will elaborate on the four pillars of modern AI systems: the data on which they rely, the task of organizing these data, the automation of this task, and the type of explanations needed to support this automation. Due to the wide span of the topics covered, the discussion is necessarily kept at a somewhat abstract level, and some prior familiarity of the reader with the field of AI is assumed.

# 1. Data Sources

Our characterization of AI as a discipline dealing with data is not related to its scientific and empirical nature. Much like any scientific and empirical discipline, AI seeks to understand a natural phenomenon — intelligence, in this case, as it predominantly relates to humans — through observations, formulation of theories and hypotheses, and their experimental evaluation and subsequent revision. But it is not the data that come in the form of observations (that drive the formulations of theories and hypotheses about the mechanisms of intelligence) that makes AI a discipline about data, but rather the fact that the formulated theories and hypotheses themselves are about mechanisms that consume and process data.

Data is widely acknowledged to have emerged as a commodity in modern times, and this emergence is not decoupled from the tight synergistic relation of data to AI, with the latter providing the methods and tools that have, on the one hand, supported the processing needed to derive value from data, and have, on the other hand, created an even bigger demand for data. At the same time, the substantial increase in computational power and the massive sharing of data facilitated by technological advancements and certain changes in our social relations and interactions have come to catalyze this synergistic relation.

## 1.1. World Wide Web

Since its inception, the World Wide Web (WWW) has facilitated the sharing and interconnectedness of information and data. With access to the WWW permeating societies and populations across the socio-economic spectrum, the availability of information and data is increasing in terms of both its accessibility but also its quantity (even if not its quality), which has been growing at an unrelenting pace.

Data on the WWW come in a wide variety of "shapes and sizes". Personal web pages host information that individuals choose to share. Corporate web pages host information on the structure of companies, their personnel, their employment opportunities, and their vision and progress towards achieving it. Electronic shop web pages host information on a wealth of products that they sell. Blog web pages host information on the ongoing topics on which their contributors wish to comment. Archive web pages host literary works whose copyright has elapsed (or not). Social networking web pages host the social interrelations of their users. Media sharing web pages host videos or audios that capture personal moments of individuals, professionally produced films and songs, or amateur live streaming of shows and concerts. Large scale project web pages host information including detailed maps of the entire Earth[1], global rankings of tertiary educational institutions[2], or notable sequences of integer numbers[3].

Other web pages act not as hosts of original content, but as aggregators of content found on other web pages. Yet other web pages allow multiple contributors to share information and exchange thoughts, supporting the collaborative creation of content[4], engagement in science[5], or solving of math problems[6].

A unique aspect of the data found on the WWW is their linked nature. Links between data may show relatedness, endorsement, or juxtaposition, either between the data themselves, or between their creators, hosts, or aggregators. Links are, then, another type of data that can be found on the WWW. Interestingly, the content-related data can act as meta-data for the link-related data, since they can specify the nature of the relation of the linked web pages. A link from a web page opining on the results of a political event towards a web page hosting a video recording of that event can be seen as endorsing or opposing the views expressed in the video depending on the data found on the opinion web page.

Although one could attempt to view the WWW as a repository of our human common knowledge, the veracity of the information found on the WWW should be far from being taken for granted. As an open, and largely unmoderated, space for sharing information, the WWW has its own demons to fight. For every news outlet that shares its reports on the WWW, there exist numerous other outlets that share opinions skewed by ideology, bias, or preconceptions, or propagate presumed news stories that are completely fake or fabricated. Recognizing that the WWW is not a source of commonly accepted truths and facts, but rather a source of websense (Michael, 2013) — rife, as it may be, with cultural biases, fictional statements, expert knowledge, common misconceptions, or deliberate lies — helps guide one on how to approach this vast source of data, and how to develop techniques and tools to create value.

## 1.2. Internet of Things

The Internet, a collection of interconnected machines serving WWW content, has naturally evolved into the Internet of Things: a collection of devices serving data coming from varied types of sensors.

Some of the data available through sensors concern the monitoring of public assets. Technological solutions can easily monitor and report, among others, the daily storage levels in a water reservoir, the weekly amount of garbage collected at the city park, and the minute-by-minute number of free spots in a municipal parking place. Other data might be gathered by public agencies, but concern the monitoring of the natural environment, the archetypical case being the monitoring, by a national weather service, of temperature and humidity levels, atmospheric allergens and contaminants, wind conditions, etc.

Even situations that could be considered as potentially infringing on the rights of individuals when the latter are being monitored without their consent, have equally been in the sphere of what is (technologically feasible to be) monitored: traffic movement, the incident of crimes, cross-cultural tensions and actions of civil unrest, population density, and other such aggregate measures.

In all cases, the utilization of public money in the development of the asset being monitored or in the operation of the monitoring sensors, has led to the argument — and, increasingly, the legal obligation — that the produced data should be shared openly and freely with the public. Not only this sharing of data allows the public to be informed about the monitored assets, but it further allows for the development of services that filter, manipulate, and combine the data in a manner that creates value for others, leading to the emergence of an Internet of Services that goes beyond an Internet of Things and Sensors.

Services are not restricted to the processing of existing data, but they also support the creation of novel data coming from virtual worlds or digital twins of physical objects and processes. Ultimately, one can talk about an Internet of Data, within which one finds a virtuous cycle, where the availability of data invariably leads to the creation of even more data, giving rise to an entire ecosystem of data processing, and lowering the bar for even individuals or small groups to leave a mark on the innovation landscape, or to produce solutions that improve human efficiency[7] in coping with the tsunami of available data.

## 1.3. "τὰ ἐν οἴκῳ ἐν δήμῳ"

Along with, or because of, the technological advancements that nullified the transmission and sharing cost for data, there has also been a societal shift on the view of personal data. Paradoxically, and despite concerns and relevant regulations about protecting the privacy of individuals, increasingly more people share private matters publicly, opposing the ancient Greek proverb "τὰ ἐν οἴκῳ μὴ ἐν δήμῳ"; loosely translated: that ("τὰ") which is of the home ("ἐν οἴκῳ") is not ("μὴ") of the community ("ἐν δήμῳ").

Spearheaded by the younger generation, many people seem willing to share personal information on online and offline sites without much thought on what that data might reveal about them. It is not uncommon for people to reveal their gender, sexual orientation, race, and religion, without realizing that such information is usually not necessary for most everyday activities that they might wish to engage in. Equally common is for people to share information that could lead to identity theft, including their national ID number, their social security number, and their credit card information; not to mention publicly announcing when they are away from home on vacation, raising the danger for house theft.

Leaving aside the willing or unwilling sharing of such explicitly sensitive information, even privacy-mindful people end up sharing data that offer cues on aspects of their profile that they may wish to keep private. The sleeping patterns of a human, as recorded by their fitness tracker or smart watch, may reveal their gender. The mere fact that someone owns such a device may offer clues on their financial status. The recording of their heart rate might reveal when they are outside of their house exercising.

Online clothes shopping might reveal the size and shape of one's body, or a potential disdain in some fabrics perhaps due to a known textile allergy. Online grocery shopping might similarly reveal one's intolerance to certain dietary substances, or a potential pregnancy! Searching the WWW might reveal the presence of a disease when searching for symptoms, or the passing away of a close friend when searching for ways to cope with the stress of loss. Voice commands to a house assistant device might reveal times when one is home alone or with friends, based on the background noise. Pictures shared online can reveal one's social acquaintances, including their type of relationships with other individuals.

That is not to say that all instances of sharing data are done frivolously. In a social context, we share data as a sign of trust, or to solicit a response from the other party. When that other party is a human, the response might be a same-kind sharing of data about themselves. When, instead, we are dealing with an organization, the response might be some sort of personalization of the services it offers.

And it is exactly herein that lies a great potential for innovation. The digitization of personal data and the social relations between people has given rise to an Internet of People[8], where one's interaction with their environment and with others can be mediated by machines to create value to all those involved.

## 2. Organization

The mere availability of data does not, by itself, offer value for a reason that is becoming increasingly more prevalent as data becomes more abundant: data is the proverbial haystack, and extracting value from the data amounts to finding the proverbial needle. How does one find the proverbial needle?

There is no single way to extract value from data, nor a single set of technologies to do so. Focusing on the organization of data as the key means of extracting value, and on the use of AI tools and techniques for carrying out this organization, offers a unifying view in the context of AI-driven innovation.

### 2.1. Searching the Past

Perhaps the simplest way that data can be processed and create value is through their searching. The massive nature of data makes any naïve searching strategy simply insufficient. A lawyer searching for precedent in the law for a particular case cannot meaningfully read through all past cases. Even assigning the task to paralegals is not necessarily a solid strategy, as the speed at which the search can be carried out and the precision and recall of the results might not meet the constraints of the case. Organizing the data in a way that allows for their efficient and reliable searching, directly by the lawyer, and their presentation in a manner that is easy to consume by the lawyer, and that allows the lawyer to verify the outcome of the search, would seem to be a necessary component of the search process.

A company keeping records of its financial transactions for auditing purposes would make the auditor's life much easier if it were to organize the records in a way that would allow the auditor to effectively and efficiently identify the information that is being sought. Even organizing one's document or media files on their personal computer can be seen to be beneficial when it comes to trying to find something.

Data organization in support of the searching task might require the taking of actions at the time the data is initially stored (e.g., by storing data along with relevant meta-data information), periodically as the data is updated, or as information is gained on how the data is being typically searched (e.g., store more frequently-accessed data in a manner that is faster to retrieve), or at the time the data is accessed (e.g., by choosing the order in which to traverse the data based on the particular search query).

An equally central aspect of searching, and one that more prominently showcases the applicability of AI-driven innovation, is identifying the underlying intention of the posed search query. Autocompleting partial queries on the basis of statistical information from past searches is a widely-used technique, with well-recognized issues.[9] More elaborate solutions might include the profiling of the one posing the queries, and attempting to reinterpret the query in a personalized manner according to that profile. Thus, the query "restaurants nearby" might be reinterpreted as "vegetarian restaurants in Nicosia" if the one searching is known to be a vegetarian residing in Nicosia, or might be reinterpreted as "traditional taverns by the sea" if the one searching is known to be hosting, at the time, colleagues from abroad who are visiting Cyprus for the first time and wish to experience the local cuisine while enjoying the weather.

In all cases above, the organization of data amounts to what one would expect to be done by a competent human assistant, who either in anticipation of a future search query, or as a result of one, prepares the data and the way to interact with and view them, in a manner that would provide the best-outcome search experience (in terms of efficiency and eventual value) for the one posing the query.

Google, one of the largest providers of search services on the WWW, explicitly states their mission as being "to organize the world's information and make it universally accessible and useful".[10] While some of the ways of organizing data that are described above have been a stable hallmark of Google's services, or were gradually added as Google evolved over the years, searching is by no means the only task associated with the organization of data, nor necessarily the one more directly utilizing AI.

## 2.2. Recognizing Patterns

Thinking of data not as isolated points of interest but as glimpses of some underlying structured reality, value could be derived by their organization in a manner that would reveal that underlying reality. Epidemiologists trying to understand the outbreak of a novel virus by analyzing the clinical data of infected individuals may wish to recognize recurring patterns across that data. The patterns that could be of possible medical interest are not necessarily known by the epidemiologists, and attempting to enumerate and test all such patterns is beyond what is humanly possible, at least in the timeframe that would still be of use for the early development of a potential treatment. Appropriately organizing the data in a manner that allows for the efficient recognition of patterns in the data seems to be necessary.

A company wishing to direct incoming phone enquiries to the appropriate department would benefit by organizing the different enquiries into groups so that each group is handled in a specialized manner. Analogously, a biologist might wish to organize the DNA data of different living or fossil organisms in a way that would reveal potential patterns in their interrelations, patterns in their relative position in a phylogenetic tree, or patterns in their scientific classification in a hierarchical taxonomic structure.

Beyond finding patterns or regularities in data, equally central is the task of identifying abnormalities in data. A network administrator monitoring the traffic in a computer network is unlikely to be able to monitor traffic in real time for potential problems. But even if this live monitoring of data could be done, it is unclear what information should make the network administrator suspect that something is possibly wrong. Ultimately, an abnormality could be considered to be anything that breaks the usual regularities in the data. And since regularities depend on the entire set of data, they cannot be known up front. Consequently, abnormalities cannot be expected to have a particular anticipated pattern. Only through the appropriate organization of data, including their proper visualization or communication to the network administrator, can abnormalities be made to stand out and be effectively recognized.

## 2.3. Making Predictions

The underlying reality revealed by data might often be one that persists into spatially or temporally novel circumstances. When identifying that there is some correlation between the cars that past loan applicants drive and whether they defaulted on their loan payments, that regularity can be reasonably assumed to persist to future loan applicants. When identifying that smoking is correlated with lung diseases in a certain country, that regularity can be reasonably assumed to persist to other countries.

Making predictions in these novel situations requires the organization of the data in a way that helps expose their regularities and maps them in a form that can be readily applied in the novel situations. In particular, the regularities should be organized in a manner that acknowledges that in the novel situations the data available might be incomplete, and that one might wish to predict the missing parts of the data by using those regularities as a guide; i.e., predict missing parts in the novel situation in a way that would make the imputed data obey the regularities that resulted from the organized past data.

Such ways of organizing data are often associated with the notion of learning, in the sense that one uses part of the data to extrapolate on what should be the case in another part of the data. Considering the familiar setting of student learning, we can see that learning is facilitated by the way the teacher organizes the data, either in a manner that provides the student explicitly with the regularities that the latter is supposed to identify, or by greatly aiding the latter in identifying those regularities through the gradual and orchestrated presentation of a particular sequence of data points to the student. At the other extreme, the teacher might simply present the data in no particular order, leaving the task of learning, i.e., that of organizing the data for the purpose of making predictions, entirely to the student.

Beyond the object-level use of learning to facilitate the making of predictions, one could also consider the use of learning at a meta-level, as a process that operates on the meta-level data determined by the interaction of an agent with the object-level data. In the context of searching past data, the posed search queries can themselves be viewed as meta-level data, and one could seek to anticipate future search queries by extrapolating from past search queries. Analogously, in the context of recognizing patterns or identifying abnormalities over data, any patterns or abnormalities that were previously accepted as being of value to an agent can be viewed as meta-level data, and one could seek to anticipate patterns or abnormalities that might be of value in the future by extrapolating from the past.

Other than organizing the data so that a lawyer, an auditor, or a computer user can quickly perform searches, the data could also be organized in a manner that recognizes the types of search queries that are usually posed in different contexts, so that the type, and even the timing, of future search queries can be predicted and anticipated, and so that the execution of those particular searches is optimized.

An epidemiologist wishing to recognize infection patterns across a population, a company wishing to recognize patterns across its received user support requests, and a network administrator wishing to identify abnormalities across its network traffic logs would further benefit by having their patterns or abnormalities being organized so that their value could be predicted and anticipated during the next pandemic, during the next deployment of a call center, or during the next designing of a network.

Given the central role that learning and prediction are able to play in the way data can be most usefully organized, both at the object-level and at a meta-level, and offer value, it is unsurprising that learning has come to be the armada in the fleet of technologies that characterize modern AI-driven innovation.

## 3. Automation

The mental or cognitive task of organizing data has, for millennia, been undertaken predominantly by human workers, with the occasional help from simple external memory aids (such as the papyrus and later the paper) and computational aids (such as the abacus and later the calculator). It is, after all, an integral part of the human endeavor to deal and cope with the regularities and abnormalities of everyday life, to be able to search the past, recognize patterns, and make predictions (Michael, 2015b).

Such inherent is, in fact, the process of organizing data in humans that even our genetic makeup is able, through natural selection, to develop and test new data organization processes and carry from generation to generation those processes found to be fruitful in previous generations (Valiant, 2009). The presence of emotions in humans is one of the organization processes that evolution has devised. Organizing certain combinations of biological signals under the group "fear" helped ancient humans to run away from an attacking beast, and helps modern stockbrokers to run away from a failing company.

With the amount of data being produced[11] per second exceeding what the senses of a typical human could possibly consume[12] in an entire year — which, itself, is nowhere close to what a human could consciously process — neither the human mind nor the cross-generational process of biological evolution can meaningfully be expected to organize these data. This excess offer of data as an untapped resource has brought to the surface a socio-economic opportunity for data organization, and the subsequent market demand for developing automated mechanisms to materialize that opportunity.

### 3.1. Industrial Revolutions

The demand for automation in data organization can be appreciated in relation and in juxtaposition to the demand for automation in the manufacturing industry. With a heavy dose of generalization and abstraction, as needed to draw lessons on how the evolution of the manufacturing industry can inspire and guide data organization, each of the industrial revolutions can be characterized along two axes: the primary capacity that facilitated its emergence, and the primary consequent that it brought about.

The First Industrial Revolution (circa 250 years ago) was primarily facilitated by the advent of the steam engine, and gave rise to machines that provided human workers with a helping hand, offloading some of the physical burden required for the manufacturing process. Unlike the technological nature of the capacity that facilitated the First Industrial Revolution, the Second Industrial Revolution (circa 150 years ago) was facilitated by a conceptual capacity: the idea of using an assembly line. Perhaps unsurprisingly, such a non-technological shift had also a non-technological social consequent: the role of humans was diminished from that of key decision-makers in this human-machine partnership into that of blue-collar workers, completing (mechanically) over the assembly line those steps of the manufacturing process that could not be undertaken by the machine itself (playing, in a sense, the role of a "biological gear" in the machine). The Third Industrial Revolution (circa 50 years ago) that followed was again driven by a technological capacity: the advent of electricity, which facilitated the further automation of manufacturing by replacing the steam engine as the underlying supporting technology. Humans continued to play a secondary role, with the majority of the manufacturing process being undertaken by machines, but now in a more efficient and global manner. The social impact came as a result of this more global automation, without, necessarily, changing the way humans and machines interacted.

The most recent and ongoing Fourth Industrial Revolution, reminiscent of the socially-driven capacity that brought about the Second Industrial Revolution, came to change again the way humans interact with machines, being guided by the socio-technical capacity of increased communication. In an era of tight interconnectedness and increased flow of information in human-to-human, machine-to-machine, and human-to-machine interactions, we are currently witnessing a new role for both machines and humans, with humans regaining their earlier social role in this human-machine partnership (which was lost during the Second Industrial Revolution) of white-collar supervisors, guiding the machines on how to move from automated manufacturing to autonomous (in terms of decision-making) manufacturing.

### 3.2. AI (& ML) Revolutions

Artificial Intelligence has emerged as a formal scientific discipline around 60 years ago, and its sub-discipline of Machine Learning (ML) — the study and practice of automating the extraction of patterns and generalizing from data — saw its first systematic push around 40 years ago. This First AI Revolution was facilitated by the scientific advancement, or the desire thereof, to understand and replicate human intelligence in machines. Although coinciding, timewise, with the Third Industrial Revolution, the primary consequent of the First AI Revolution was analogous to that of the First Industrial Revolution in that it helped offload some of the cognitive burden required for certain data organization tasks.

Despite its short history, AI is currently undergoing its Second Revolution, coinciding, timewise, with the Fourth Industrial Revolution, but being, in terms of its primary facilitator and consequent, closer in spirit to the Second Industrial Revolution. This Second AI Revolution is primarily facilitated, at least in the context of using AI for innovation purposes, by the conceptual capacity brought about by Deep Learning: the scientific and engineering sub-discipline of Machine Learning that obviates the need for the careful human curation of data before those are fed into the ML pipeline, leaving, instead, ML itself to identify and extract, implicitly, those features of the data that are deemed to be useful for learning.

Not unlike the Second Industrial Revolution, then, the primary consequent of the Second AI Revolution is a social one: the role of humans was diminished from that of the key data curators in this human-machine partnership into that of a blue-collar workers, completing (mechanically) over the "assembly line" those steps of the learning process that could not be undertaken by the machine itself: namely, the labeling of the data with the expected extra-data inferences that the machine would try to draw.

The subsidiary role that humans have taken to AI-driven machines as part of the Second AI Revolution has an admittedly much wider social impact than the analogous change in role that occurred as part of the Second Industrial Revolution. Automation in the manufacturing industry had primarily meant the cession to machines of some of the more mundane (and primarily physical) forms of control that humans had exercised that far. Even under a presumed "worst-case" scenario where automated machines in a manufacturing plant were to "go haywire", the resulting consequences would be of a limited and local scope, resulting, perhaps, in a bad batch of products and the loss of some income.

AI-driven automation, on the other hand, can already be seen to have planet-wide consequences in case of a malfunction. A huge number of decisions that determine the individual socio-economic status of billions of people are taken daily with the help of AI-driven technologies: determining the granting of loans, the administration of health services, or the admissibility of evidence in criminal cases. The consequences can easily be seen to be even more cataclysmic when considering the use of AI in stock exchange transactions, in warfare decisions, and in the evaluation and handling of global health crises.

### 3.3. Ceding Control to AI

The concern of losing control to Artificial Intelligence was foreseeable from the early days of AI, even finding its way into popular films and books of the day. With that concern becoming more tangible and its effects more consequential during the ongoing Second AI Revolution, all sectors of the quadruple helix have increasingly become involved in trying to understand, regulate, and bound how much control, in which contexts, and at what cost, should be endowed to AI-driven technologies and AI systems.

With the Second AI Revolution still in its early years, the effort of tackling the concern of losing control to AI is under way. Yet, certain, even if not completely unambiguous and fully fleshed-out, lines of attack have emerged, and concern the need for fairness, accountability, and transparency in AI-driven technologies. To some extent, all these requirements are centered on a common underlying position: the expectation for some form of guarantees on the operation of the machine to which control is ceded. This high-level position remains applicable also in the context of the automated organization of data.

Fairness, then, is the requirement that similar data are organized in a similar manner. When searching for images of doctors on the WWW, fairness could amount to receiving images of both male and female doctors, perhaps in a ratio that reflects the actual statistics of gender in that profession. When seeking to recognize patterns in past litigation cases, fairness could amount to not considering the race of the judge to be relevant. When making predictions on an applicant's future loan defaults, fairness could amount to not using the applicant's religious beliefs as a discriminating factor for the decision. Not all cases of fairness have to do with discrimination. Fairness would also be to rank higher a candidate with a higher entrance exam score, to group together works of art that are of similar aesthetical value, etc.

Accountability is the requirement to make precise the bearer of the responsibility and the cost for the consequences that result from the organization of the data in each context. A machine that classifies a sick person as being healthy and advices a doctor to withhold medical treatment, or that fails to stop an autonomous car at a crossroad that causes in an accident, would seem to bear some responsibility. Since responsibility in our societies is still reserved, legally or ethically, for human actors, who should inherit that responsibility? Should it be the machine's owner, manufacturer, or trainer? And how does the particular context each time affect the determination of the answer to this preceding question?

Transparency aims to facilitate data organization in a manner that promotes fairness and accountability, or at least in a manner that helps determine when these other requirements are not met. Rather than being a functional requirement on how data are organized, transparency imposes restrictions on the structure of the machine that organizes the data. This structural requirement effectively allows an auditor to examine what led to a particular decision by an AI-driven machine, and to identify the reasons behind something going wrong. More generally, transparency allows the auditor to interact meaningfully with a machine even without having a prior opinion of what is acceptable in each possible context, but also in cases where the auditor is willing to be told and convinced by the machine.

In the case of making recommendations to a traveler on which hotel to book, a transparent machine would expose to the traveler the reasons in support of the machine's recommendation, or more generally, reasons in support of various potential options, leaving the traveler to decide how to proceed.

### 3.4. Fourth AI Revolution

The transparency requirement for AI-driven technologies is, ultimately, a way to gain back some of the control ceded by a human to a machine, as it allows the human to have an informed opinion on whether the machine is acting appropriately (according to the particular human) in each given situation.

In principle, this informed opinion can be obtained without knowledge of an AI system's internal processes. A system could be opaque on how it reaches each particular decision, but it could allow for its decisions to be contested and checked against predetermined specifications (Tubella, et al., 2020). This would be sufficient for determining whether the machine acted appropriately, yet it would offer little in terms of building trust towards the machine acting appropriately in future (uncontested) cases.

Instead, a more dialectical interaction with machines able to explain and be explained to, in a manner that is cognitively-compatible with the particular humans with which the machines will interact, is expected to be catalytic in building long-term trust and offering assurances on their future behavior.

The direction foreshadowed by the now considerable discussion among the relevant stakeholders on the need for explainable AI systems is that of an oncoming Fourth AI Revolution, analogous in nature to the Fourth Industrial Revolution.[*] Facilitated by the increased (and need for even more) communication between humans and machines, the Fourth AI Revolution is expected to turn automated AI-driven technologies into autonomous ones, with their autonomy being guided, evaluated, and monitored through the exchange of explanations, ensuring that they respect, at least, certain basic human values.

## 4. Explainability

The type of automation that would meet the requirements for machines in the Fourth AI Revolution can perhaps be summarized in two, seemingly contradictory, aims: super-human abilities in organizing the data, yet human-level abilities in communicating with human coworkers or decision-makers. On the one hand, machines are expected to continue doing what they are good at: sifting through massive amounts of data exploiting super-human computational resources, memory capacity, and parallelization. On the other hand, machines are expected to also do what humans are good at: being sociable, interacting and working with — not only for — others, and adapting to the particularities of individual coworkers.

This sought dual capacity of explainable automation fits squarely within the scope and vision of the Fourth AI Revolution, and with the current emphasis of academic and industrial research on human-centered AI, human-in-the-loop approaches, and a future of human-machine symbiosis rather than only of human-machine interaction. The fact that explainability has come to be considered a key feature of such a symbiotic relation should not be a surprise. Explanations of decisions are central in human-to-human interactions, and form the bedrock of several modern social institutions and practices.

When evaluating whether an action is ethically justified, we do so not merely in terms of the action itself out of any context, but in terms of the intentions of the actor and the current context, which we expect to be unveiled by asking the actor to explain themselves. Even if we take a consequentialist view of ethics (i.e., "the end justifies the means"), we still wish to understand what end the actor had in mind to justify their decision and action. Analogously, the role of an explanation within our legal systems (perhaps in the form of an apologia in front of a judge) plays a central role, if not in determining whether a crime has been committed, then in determining the punishment that is fitting for the criminal.

Automation that comes through an ethically-responsible machine is bound to have explainability at its core, tightly integrated into its very design, and not as a post-deployment afterthought (Michael, 2020). Importantly, explainability should go beyond the simplistic offering of one-way post-hoc explanations, and acknowledge that in a symbiosis, bilateral and real-time explanations are essential (Michael, 2019).

### 4.1. Legal Considerations

---

[*] The primary facilitators (i.e., electricity) and consequences (i.e., automation) that would correspond to a Third AI Revolution, in analogy to the Third Industrial Revolution, have been a stable feature of the first two AI Revolutions. We skip forward to the Fourth AI Revolution to emphasize the analogy with the Fourth Industrial Revolution.

The urgency of the need for AI-driven technologies not to be left unchecked, and the maturity of the discussion surrounding this need, is reflected in the actions of stakeholders across the quadruple helix.

The tight interconnectedness of the four sectors of the quadruple helix makes it hard, and perhaps unnecessary, to identify which sector spearheaded this discussion and when. But, for the purposes of exposition, one would not be far off to attribute early discussions to the public sector (including in literary works) and to the fear of autonomous robots creating a dystopian society. As technology and AI progressed, these concerns were voiced by prominent members of the public[13], and found their way into the academic sector, which for several years now has included special themes, workshops, and panel discussions in its premier scientific conferences and journals that explicitly tackle the ethical dimension surrounding the use of AI, and more recently the need for explainable human-centered AI.[14]

Either in response to the heated discussion happening in the public and academic sectors — following the maxim "Vox Populi, Vox Dei" — or in anticipation of a potential abuse of AI-driven technologies and their growing and unconstrained use across application domains, governments took a step in setting up legal frameworks to regulate the use of AI. Perhaps the most prominent and mature such effort is European Union's General Data Protection Regulation (GDPR)[15], which came into effect in mid-2018.

This is not to suggest that members of the industry sector have not played an important role in pushing the discussion forward. But the contribution of the industry sector has certainly been more instrumental in demonstrating a wide range of thought-provoking applications of AI-driven technologies, making clear the impact that these technologies can have on the rights and freedoms of individuals and the society.

GDPR includes numerous provisions on how data should be stored, organized, and accessed, aiming to preserve what has come to be a central human right in modern societies: the right to privacy and control of one's personal data. On the other hand, GDPR also includes explicit provisions on the automated processing of personal data in terms of the rights to: request access to that data, request an explanation on decisions reached through the processing of data, refuse or challenge automated decisions, etc.

Although the right to challenge, or contest (Tubella, et al., 2020), an automated decision is distinct from (and considerably less-studied than) the right to explanation, it is not completely independent. Having a process to contest a decision, including towards determining the ethical basis for that decision, can be meaningful only if the interlocutors can exchange explanations that are cognitively-comprehensible to the other party, through a process that resembles an argumentative dialogue (Kakas & Michael, 2020).

### 4.2. Post-hoc vs By Design

The predominant, even if not the only, way in which AI has supported the development of innovative solutions through the automated organization of data is through the use of Deep Learning (DL). The very design of DL-based architectures has led to their characterization as black boxes in that they are opaque on how the learning process that they support organizes the data, and, by extension, on how the resulting organization of data leads to the making of predictions or the recognition of patterns.

The initial reaction of researchers and practitioners to the need for explainability of existing AI systems was, understandably, the retention of the power of DL techniques and their "patching up" or "upgrading" with some ability to explain their decisions after those are derived. A prominent approach to this post-hoc form of explainability comes by using a pre-trained DL model to train a twin white box model, such as a decision tree classifier. Whenever an explanation is required for the decision of the DL model, one effectively requests the twin model to offer that explanation in the stead of the DL model.

On first reading, this approach might seem to enjoy the best of both worlds: the ability of DL to construct models from data exhibiting a complex structure, and the ability of white box models to offer explanations. The situation is, however, more nuanced than this first reading. The explanations that result from this approach do not really explain the mechanism by which the predictions were derived in the first place. Nor do these explanations offer a reason of why the predictions should be accepted by a human. Rather, the explanations capture certain conditions that are expressible in the specific syntax of the white box models, and that happen to hold in the data that gave rise to the predictions.

A second line of attack to the explainability problem, then, is to devise learning algorithms and learning representations that by design ensure that the learned model will be able to offer direct explanations of why a prediction was made, or more pertinently, of why a prediction should be accepted by a human.

Explainability by design seems to be closely linked to symbolic representations of learned models. Such representations were thought to be of key importance in the development of AI systems during the early days of AI as a scientific field, and have remained popular among the AI sub-community that has continued working on the problem of representing and reasoning with knowledge. They have, however, become increasingly disconnected from the AI sub-community that focused on the problem of learning knowledge, and which nowadays includes those using DL techniques for that problem. In a sense, then, the emergence of explainability as a key consideration in the development of AI systems has pointed back to some of the early ideas, and has given a breath of fresh air to knowledge-based AI techniques.

### 4.3. Universal vs Contextual

Orthogonal to the question of what representations need to be in place for an AI system to most effectively support explainability, is the question of who is the recipient of the derived explanations.

An operating system suggesting to an ordinary computer user to save their work and attempt to reboot their machine might reasonably provide an explanation that "a system application has malfunctioned". Such an explanation would, however, be insufficient to a software developer being paid to ensure that the operating system works as expected, whereas the more technically-savvy explanation "application 'system.exe' has thrown an exception: more than 40 documents open" would certainly be more useful.

What manifests as technical depth in the explanations of the scenario above is, more generally, a notion of cognitive compatibility of the explanations with their intended consumer. The explanations that are comprehensible to one human may differ significantly from those that are comprehensible to another.

Even among comprehensible explanations, the broader context in which explanations are given might make certain explanations more appropriate than others. The fact that milk is a dairy product and the fact that milk contains lactose are both comprehensible explanations for the suggestion to a human to avoid drinking milk, and could be interchangeable in certain settings. Yet, the former is clearly more appropriate if the human is vegan, and the latter is more appropriate if the human is lactose-intolerant. Both explanations would likely be irrelevant to a non-vegan and non-lactose-intolerant individual.

The milk scenario above hints, in fact, to another important dimension of the contextual nature of explanations: their contrastive nature (Miller, 2019). Explanations are meant not only to support the taking of a certain decision by the machine, but also to support the non-taking of an alternative one. Which among the numerous alternative decisions is to be contrasted against has no simple answer.

### 4.4. Unilateral vs Bilateral

The natural flow of explanations, as a way to support transparency and take back some of the control that humans have ceded to AI systems, is from a machine to a human. The machine makes a decision or performs an action, and it then offers an explanation to a human in support of that choice. A question that arises effortlessly is whether such unilateral explainability from a machine to a human is sufficient.

On the one hand, it is. The machine is providing the human — who is attempting to be in control, or at least in the loop — with the necessary information needed for the human to understand and evaluate the machine's choices. In a sense, that is what all machines do, be it through visual or auditory cues (e.g., a car sounding a proximity alert just before it auto-engages the breaks to reduce the impact speed), or through the offering of a natural language explanation (e.g., a computer informing the user that it is about to turn off due to a low battery, and it is auto-saving and closing all open documents).

And this would have been the end of the answer to the first question if it were not for another related question. How does a machine become able to offer explanations by design that are contextual in nature? Machines acquire skills primarily in one of two ways: through programming or through learning. Either someone pre-programs a machine on what explanation is appropriate in each particular context, or someone trains the machine on example contexts by providing an appropriate explanation in each one. In either case, the human interacting with the machine ends up offering explanations to the latter.

A principle from Computer Science states that the quality of the output of a machine depends on the quality of the input, often described as "Garbage In, Garbage Out (GIGO)". One could adopt an analogous stance for explainability: the ability of a machine to output explanations depends on its ability to receive explanations, which can be described as "eXplanations In, eXplanations Out (XIXO)". This stance suggests, then, that bilateral explainability is not only useful, but also necessary (Michael, 2019).

Through the ability of a machine to receive explanations, the machine acknowledges not only that the "correct" explanation depends on the context, but also that the context includes the human with whom the machine interacts. In the same way that different humans would expect personalized explanations from the machine, so they, in turn, will offer personalized explanations to the machine. The machine will, consequently, be prepared to offer explanations that are in line with what each human expects.

A somewhat subtle implication of adopting the XIXO principle is that the incoming explanations should be produced by whoever will be consuming the outgoing explanations. This, in turn, implies that contextual explainable AI systems cannot be fully developed a priori, but that they have to be trained in-situ and on the job, while interacting with their intended user; not by being prepared in the lab.

## 4.5. Real-Time vs Delayed

The need for an in-situ interaction between AI systems and humans does not, prima facie, preclude the interaction from occurring sometime after, and not during, the event that necessitates the interaction.

An autonomous car suddenly swerving and narrowly missing a bicyclist that was right in front of the car is a scenario that might necessitate an interaction. The car passengers might find the action reasonable, perhaps by projecting their own intentionality and believing that the car swerved to avoid hitting the bicyclist. If, indeed, this is the explanation that the car will offer in support of its decision when it ends up interacting with the passengers, then the passengers would find it acceptable, in the sense that an analogous decision would be a reasonable one to be made in analogous future situations. If, however, the car offers the explanation that exactly one year has elapsed from the car's manufacturing, and the car swerved as a way to commemorate its "birthday" event, then the act of missing the bicyclist might be considered a fortunate side-effect, while the reasoning behind the decision might be unacceptable.

An automated image-cropping system cropping a picture[16] from a class outing so that it excludes a black girl standing at the sidelines of another group of students is another scenario that might necessitate an interaction. A casual viewer of the picture might find the action objectionable, potentially ascribing gender or racial bias to the system. If, indeed, the system ends up explaining its decision to exclude the girl from the picture simply in terms of her gender or race, then the reasoning for the system's choice would reinforce the ascription of bias. If, however, the system offers the explanation that this is a class outing for a boys-only school, and the pictures were taken to be included in the school's yearbook, then the offered reasoning might make the decision more palatable to the viewer of the cropped picture.

In both scenarios, the evaluation of the decision and its accompanying explanation happen after the fact, when the decision has already been acted upon and has brought about change in the environment. Depending on the particularities of the situation, this delayed explanation might be sufficient. In the first scenario above, the effects were positive even though the explanation was inappropriate for the decision. But, in a sense, no harm was done, and the human might seek to improve the machine's future decision-making. In the second scenario above, the effects were perceived as negative even though the explanation eventually justified the decision. Again, there is no apparent harm, since the perceived bias was not really there. Should we, then, be content with receiving explanations in such a delayed manner?

Well, no! For one, there are scenarios where neither the effects nor the explanation of a decision are acceptable. And in those cases, harm might clearly occur, without an opportunity for early intervention. We would not want a medical system to prescribe strong antidepressant drugs to a patient that came in with flu symptoms, by justifying its decision on the observation that the patient did not smile at all during their medical examination, and have those drugs end up causing suicidal thoughts to the patient.

Even if we restrict attention to the two aforementioned scenarios, we would be amiss not to consider a potentially more subtle issue. In the image-cropping scenario there is a clear danger of a human evaluator being pressed, perhaps unconsciously so, to find the provided explanation to be a sufficient justification, just because they would not want to admit to themselves that a biased decision was made.

It would seem that a cleaner, although admittedly not always applicable, solution would be to have the machine act as a consultant or assistant to the human, proposing decisions along with explanations, and letting the human vet the decisions before any action is taken (either by the machine or the human).

### 4.6. Deciding vs Supporting

Shifting from a decision-making to a decision-supporting perspective raises an intriguing prospect: the machine can entertain multiple potential decisions and supplement each with an explanation. The machine is, then, proposing alternatives among which the human will choose, and offering arguments in support of, or against, each alternative to help the human make that choice (Kakas & Michael, 2020).

An AI system tasked to help a civil servant to decide on the frequency with which to send the garbage collection truck through a particular neighborhood might present the following alternatives: keep the frequency to twice a week, because that is the default choice for all neighborhoods; increase the frequency to three times a week, because that neighborhood has been consistently producing more garbage than what is typical for other neighborhoods; reduce the frequency to once a week, because that neighborhood has been consistently delinquent in settling up their garbage collection fee debt. None of the alternatives is acted upon, nor is any one treated preferentially. Rather, the civil servant bears the responsibility of making the decision, by weighing the arguments offered by the machine.

Does that imply that the control is now fully back to the human, without the active involvement of the machine? Not at all. The choice of the alternatives and the choice of the associated explanations is still a decision that is made by the machine without the intervention of the human, and it does affect what the ultimate choice of the human will be. If the last alternative in the garbage collection scenario was not offered or it was offered and was associated with a weaker argument, then the human decision-maker might not consider that choice at all, or reject it in the presence of the other alternatives.

A more pertinent view of the situation — in line with the vision of the Fourth AI Revolution — is that the human and the machine are making decisions as a team, each contributing to the team according to their own competencies. Following the XIXO principle, the human and the machine bilaterally exchange explanations / arguments to reach their joint decision, and to improve their future joint performance.

Recognizing that this joint performance, and not who makes the decisions, is the ultimate metric of success brings about a further consideration. The machine's explanation associated with the suggestion to increase the frequency of garbage collection is, ultimately, an explicit prediction that in the future the neighborhood will (continue to) produce more garbage than what is typical for other neighborhoods. Even if this statement were not accurate at the time that the explanation was offered, the very offering of the explanation might initiate a chain of events that make the prediction a self-fulfilling prophecy (e.g., seeing that garbage is collected three times a week, the neighborhood starts producing even more garbage); or that cause an otherwise correct prediction to be falsified (e.g., annoyed by the increased traffic caused by the garbage collection truck, the neighborhood stops producing excess garbage).

Properly taking into account the future ramifications of the very offering of (predictive) explanations — through a process aptly called introspective forecasting (Michael, 2015a) — even when machines act as decision-supporters, remains an under-acknowledged issue with major implications (Michael, 2015b).

### 4.7. Organization Catalyst

Over and above the use of explainability as a mechanism for restricting or qualifying the way data organization is automated, the insistence on building explainable automated systems has, in fact, the further consequence of catalyzing the systems' ability to organize data more efficiently and effectively.

Explanations transmitted from a human to a machine — as necessitated by the adoption of the XIXO principle — encapsulate in a concise manner multiple supervision signals for the training of the machine. The single explanation "You should have blocked the incoming call since I was in a work meeting." from a human to their AI-powered call-handling assistant proactively labels all future instances of incoming calls during work meetings with the decision "block", without requiring these instances to actually materialize, training the machine beyond what traditional supervised learning can offer (Michael, 2019).

Other than the resource gains in terms of time and data needed for learning, having humans associate their feedback with explicit explanations can be argued to engage the mind's logical and effortful (System 2) reasoning process (Kahneman, 2011), which may lead humans to being more consistent on the supervision signals that they offer across their interactions with the machines, potentially even reducing any unconscious biases and misconceptions that the humans might transmit to the machines.

On the other side, explanations transmitted from a machine to a human provide insights into why certain machine decisions are wrong, helping humans efficiently debug the machine knowledge. This is particularly applicable in cases where the training data on which the machine is prepared might come from not fully reliable resources (e.g., the WWW), requiring the use of an ex post curation process.

More broadly, the advent of explainability as a central-stage concept in AI is catalyzing the adoption of AI-driven solutions for innovation by facilitating the powerful integration of modern AI technologies with the more traditional symbolic self of AI (Kakas & Michael, 2020) (Tsamoura, et al., 2021), and by suggesting a future away from requiring AI systems to have "too much intelligence" and deep analytical abilities as a prerequisite for being successfully deployed in the market, and towards accepting AI systems that have "more down to earth intelligence" and that produce potentially sub-optimal but still satisficing solutions. And this future is fulfilling the collaborative promise of the Fourth AI Revolution, reassuring the average human that AI systems do not have to be approached as the opaque and distant overlords that emerged from the Second AI Revolution, but as transparent and sociable coworkers.

# Bibliography

Kahneman, D., 2011. *Thinking, Fast and Slow.* New York: Farrar, Straus and Giroux.

Kakas, A. & Michael, L., 2020. Abduction and Argumentation for Explainable Machine Learning: A Position Survey. *arXiv:2010.12896.*

Kruger, J. & Dunning, D., 1999. Unskilled and Unaware of It: How Difficulties in Recognizing One's Own Incompetence Lead to Inflated Self-Assessments. *Journal of Personality and Social Psychology,* 77(6), p. 1121–1134.

Manyika, J. et al., 2013. Disruptive Technologies: Advances That Will Transform Life, Business, and the Global Economy. *McKinsey Global Institute (http://www.mckinsey.com).*

McCarthy, J., Minsky, M., Rochester, N. & Shannon, C., 1955. A Proposal for the Dartmouth Summer Research Project on Artificial Intelligence.

Michael, L., 2013. Machines with WebSense. *Proceedings of the 11th International Symposium on Logical Formalizations of Commonsense Reasoning.*

Michael, L., 2015a. Introspective Forecasting. *Proceedings of the 24th International Joint Conference on Artificial Intelligence,* p. 3714–3720.

Michael, L., 2015b. The Disembodied Predictor Stance. *Pattern Recognition Letters,* Volume 64, p. 21–29.

Michael, L., 2019. Machine Coaching. *Proceedings of the IJCAI 2019 Workshop on Explainable Artificial Intelligence,* p. 80–86.

Michael, L., 2020. Machine Ethics through Machine Coaching. *Proceedings of the 2nd Workshop on Implementing Machine Ethics.*

Miller, T., 2019. Explanation in Artificial Intelligence: Insights from the Social Sciences. *Artificial Intelligence,* Volume 267, p. 1–38.

Tsamoura, E., Hospedales, T. & Michael, L., 2021. Neural-Symbolic Integration: A Compositional Perspective. *Proceedings of the 35th AAAI Conference on Artificial Intelligence,* p. 5051–5060.

Tubella, A., Theodorou, A., Dignum, V. & Michael, L., 2020. Contestable Black Boxes. *Proceedings of the 4th International Joint Conference on Rules and Reasoning,* Volume LNCS 12173, p. 159–167.

Valiant, L., 2009. Evolvability. *Journal of the ACM,* 56(1), p. 3:1–3:21.

[1] https://www.google.com/maps/

[2] https://www.timeshighereducation.com/world-university-rankings/

[3] https://oeis.org/

[4] https://www.wikipedia.org/

[5] https://www.zooniverse.org/

[6] https://polymathprojects.org/

[7] https://ifttt.com/

[8] https://www.internetofus.eu

[9] https://www.theguardian.com/technology/2016/dec/04/google-democracy-truth-internet-search-facebook/

[10] https://about.google/

[11] https://www.forbes.com/sites/bernardmarr/2018/05/21/how-much-data-do-we-create-every-day-the-mind-blowing-stats-everyone-should-read/

[12] https://www.britannica.com/science/information-theory/Physiology/

[13] https://www.bbc.com/news/technology-30290540/

[14] https://www.journals.elsevier.com/artificial-intelligence/call-for-papers/special-issue-on-explainable-artificial-intelligence/

[15] https://eur-lex.europa.eu/eli/reg/2016/679/oj/

[16] https://blog.twitter.com/en_us/topics/product/2020/transparency-image-cropping.html